\documentclass{article}

\usepackage{microtype}
\usepackage{graphicx}
\usepackage{subcaption}
\usepackage{booktabs} 

\usepackage{hyperref}

\usepackage[accepted]{icml2025}

\usepackage{amsmath}
\usepackage{amssymb}
\usepackage{mathtools}
\usepackage{amsthm}
\usepackage{bm}
\usepackage{xcolor}
\usepackage{colortbl}
\usepackage[capitalize,noabbrev]{cleveref}

\newcommand{\ours}{MolSculpt}
\newcommand{\bound}[1]{\textcolor{gray}{#1}}
\theoremstyle{plain}

\theoremstyle{definition}

\theoremstyle{remark}

\definecolor{Green}{RGB}{52,170,82}
\usepackage[textsize=tiny]{todonotes}

\usepackage{xspace}
\usepackage{multirow}
\definecolor{codegreen}{rgb}{0,0.6,0}
\definecolor{codegray}{rgb}{0.5,0.5,0.5}
\definecolor{codepurple}{rgb}{0.58,0,0.82}
\definecolor{backcolour}{rgb}{0.95,0.95,0.92}

\definecolor{mydarkred}{rgb}{0.8,0.02,0.02}
\definecolor{mydarkorange}{rgb}{0.40,0.2,0.02}
\definecolor{mypurple}{RGB}{111,0,255}
\definecolor{myred}{rgb}{1.0,0.0,0.0}
\definecolor{mygold}{rgb}{0.75,0.6,0.12}
\definecolor{mydarkgray}{rgb}{0.66, 0.66, 0.66}
\definecolor{mygray}{gray}{0.9}

\newcommand{\ignorethis}[1]{}

\makeatletter
\DeclareRobustCommand\onedot{\futurelet\@let@token\@onedot}
\def\@onedot{\ifx\@let@token.\else.\null\fi\xspace}

\makeatother

\makeatletter

\newcommand\footnoteref[1]{\protected@xdef\@thefnmark{\ref{#1}}\@footnotemark}
\makeatother

\icmltitlerunning{MolSculpt: Sculpting 3D Molecular Geometries from Chemical Syntax}

\begin{document}

\twocolumn[
\icmltitle{MolSculpt: Sculpting 3D Molecular Geometries from Chemical Syntax}



\icmlsetsymbol{equal}{*}

\begin{icmlauthorlist}
\icmlauthor{Zhanpeng Chen}{pku}
\icmlauthor{Weihao Gao}{pku}
\icmlauthor{Shunyu Wang}{pku}
\icmlauthor{Yanan Zhu}{pku,zhu}
\icmlauthor{Hong Meng}{pku}
\icmlauthor{Yuexian Zou}{pku}
\\
\end{icmlauthorlist}

\icmlaffiliation{pku}{AI for Science (AI4S)-Preferred Program, Peking University Shenzhen Graduate School, China}
\icmlaffiliation{zhu}{Faculty of Materials Science, Shenzhen MSU-BIT University, Shenzhen, China}
\icmlcorrespondingauthor{Yuexian Zou}{zouyx@pku.edu.cn}


\icmlkeywords{Machine Learning, ICML}

\vskip 0.3in
]



\printAffiliationsAndNotice{}  

\begin{abstract}
Generating precise 3D molecular geometries is crucial for drug discovery and material science. While prior efforts leverage 1D representations like SELFIES to ensure molecular validity, they fail to fully exploit the rich chemical knowledge entangled within 1D models, leading to a disconnect between 1D syntactic generation and 3D geometric realization. To bridge this gap, we propose {\ours}, a novel framework that ``sculpts" 3D molecular geometries from chemical syntax. {\ours} is built upon a frozen 1D molecular foundation model and a 3D molecular diffusion model. We introduce a set of learnable queries to extract inherent chemical knowledge from the foundation model, and a trainable projector then injects this cross-modal information into the conditioning space of the diffusion model to guide the 3D geometry generation. In this way, our model deeply integrates 1D latent chemical knowledge into the 3D generation process through end-to-end optimization. Experiments demonstrate that {\ours} achieves state-of-the-art (SOTA) performance in \textit{de novo} 3D molecule generation and conditional 3D molecule generation, showing superior 3D fidelity and stability on both the GEOM-DRUGS and QM9 datasets. Code is available at \href{https://github.com/SakuraTroyChen/MolSculpt}{https://github.com/SakuraTroyChen/MolSculpt}.
\end{abstract}
\section{Introduction}
\begin{figure}[htbp]
    \centering 
    \includegraphics[width=\linewidth]{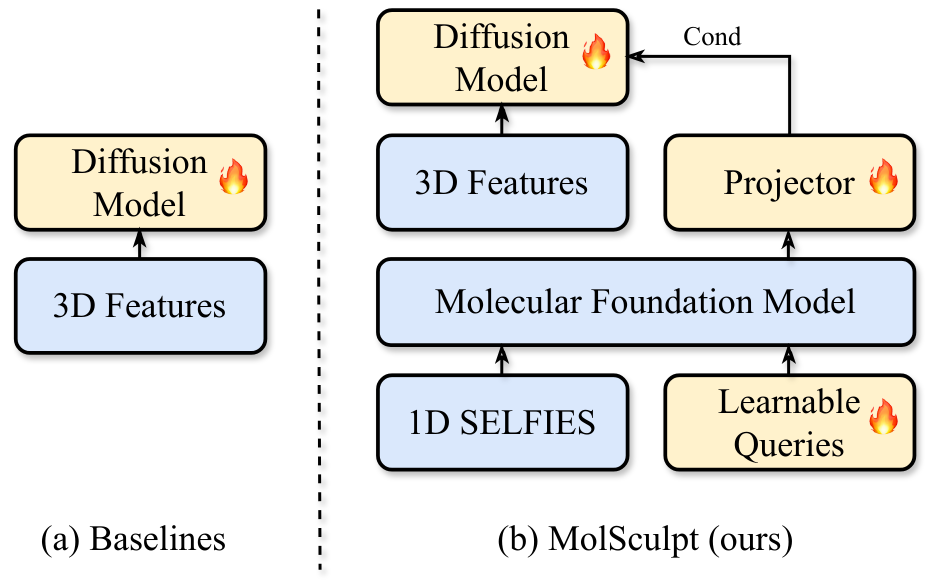} 
    \caption{Schematic illustration of the proposed {\ours}. Unlike baselines (a) that rely solely on 3D features, {\ours} ``sculpts" geometries with a cross-modal conditioning mechanism. It leverages a projector and learnable queries to transfer inherent chemical knowledge from a pre-trained foundation model to guide high-fidelity 3D molecule generation.}
    \label{fig:intro}
\end{figure}

Molecule discovery drives advances in therapeutics, catalysis, and materials by prioritizing candidates across vast chemical space, yet traditional pipelines based on hand-crafted fingerprints, physics heuristics, and string formats (e.g., SMILES) suffer from limited expressivity, poor transfer, and fragility at scale~\cite{kearnes2016molecular}. Data-driven deep learning, particularly graph-based models, learns task-specific representations end-to-end and improves prediction and controllable generation~\cite{gilmer2017neural,jin2018junction}. Representing molecules as 2D graphs (atoms as nodes, bonds as edges) with message passing captures topology and chemical context effectively, but many properties hinge on 3D geometry—conformation, sterics, electrostatics, chirality, and binding poses—motivating a shift to 3D generation for shape- and pose-aware design~\cite{riniker2015better,hoogeboom2022equivariant}. Key applications include rapid conformer ensemble generation for screening, pocket-conditioned ligand design, and diffusion-based docking for accurate, scalable pose prediction~\cite{riniker2015better,peng2022pocket2mol,corso2022diffdock}.

3D molecule generation seeks to produce chemically valid 3D conformers based on structure-aware modeling of molecules in continuous space~\cite{hoogeboom2022equivariant}. Early approaches~\cite{vignac2023midi,huang2024learning} have advanced conformer quality by directly learning equivariant, continuous 3D distributions, improving stability and adherence to valency constraints; however, they still occasionally yield invalid structures, which impedes faithful learning of chemically meaningful motifs such as pharmacophoric functional groups. To mitigate this, NExT‑Mol~\cite{liu2025next} focuses on decoupling syntactic validity from 3D geometry, which introduces MoLLaMA to predict 3D conformers from 1D string representations generated via SELFIES~\cite{krenn2020self}. Because SELFIES is intrinsically robust~\cite{polykovskiy2020molecular,krenn2020self,fang2023domain}—guaranteeing 100\% validity by construction—the pipeline ensures all generated molecules satisfy basic rules, thereby stabilizing downstream 3D generation and enhancing the reliability of learned molecular distributions. 

Despite SELFIES improving molecular validity, NExT‑Mol does not fully leverage the rich inherent chemical knowledge encoded in MoLLaMA, and current approaches~\cite{luo2025towards,song2024unified} still lack mechanisms for cross‑modal interaction and complementation of chemical information. Consequently, 1D molecule generation and 3D molecule generation operate in isolation. In the domain of unified multimodal understanding and generation, recent studies~\cite{zhang2025unified} systematically investigate whether mixed training across understanding and generation leads to mutual benefits, showing that unified VLMs trained on both task types generalize better and that alignment between input/output multimodal spaces is critical for transfer. Unified multimodal architectures~\cite{ai2025ming,ge2024seed,pan2025transfer} that route modality‑specific tokens through a language or multimodal interface and attach diffusion‑based generators demonstrate that a single model can support perception and generation across modalities, highlighting the importance of adapters that align representations for cross‑task transfer.

Inspired by these advances and challenges, we propose {\ours}, a novel architecture for sculpting 3D molecules from ``chemical syntax" (SELFIES). {\ours} builds upon a 1D foundation molecular model, MoLLaMA, and a 3D molecular diffusion model. We introduce a set of randomly initialized, learnable queries to extract inherent chemical knowledge learned during pre-training from the foundation model. To effectively transfer the latent chemical knowledge, we design a trainable projector to map this cross-modal information into the conditional space of the diffusion model and optimize the entire system end-to-end based on the original diffusion objective, thereby enhancing the effectiveness of 3D molecular generation. This progression—from direct 3D diffusion over atoms to validity-guaranteed sequence-to-3D pipelines—highlights a shift toward combining rigorous chemical validity with expressive 3D generative modeling.

In a nutshell, the main contributions of this work are as follows: 
(I) {\ours} fully utilizes the validity of 1D SELFIES and further explores the inherent chemical knowledge within 1D molecular foundation models to improve the generation of 3D molecules.
(II) {\ours} can extract 1D/2D chemical conditions from the frozen MoLLaMA and transfer the capabilities of the molecular foundation model for knowledge-augmented molecule generation, enhancing \textit{de novo} 3D molecule generation and conditional 3D molecule generation.
(III) Extensive experiments on \textit{de novo} 3D molecule generation and conditional 3D molecule generation demonstrate that {\ours} consistently improve molecular validity, stability, and accuracy. 

\section{Related Work}
\begin{figure*}[t]
    \centering
    \includegraphics[width=\linewidth]{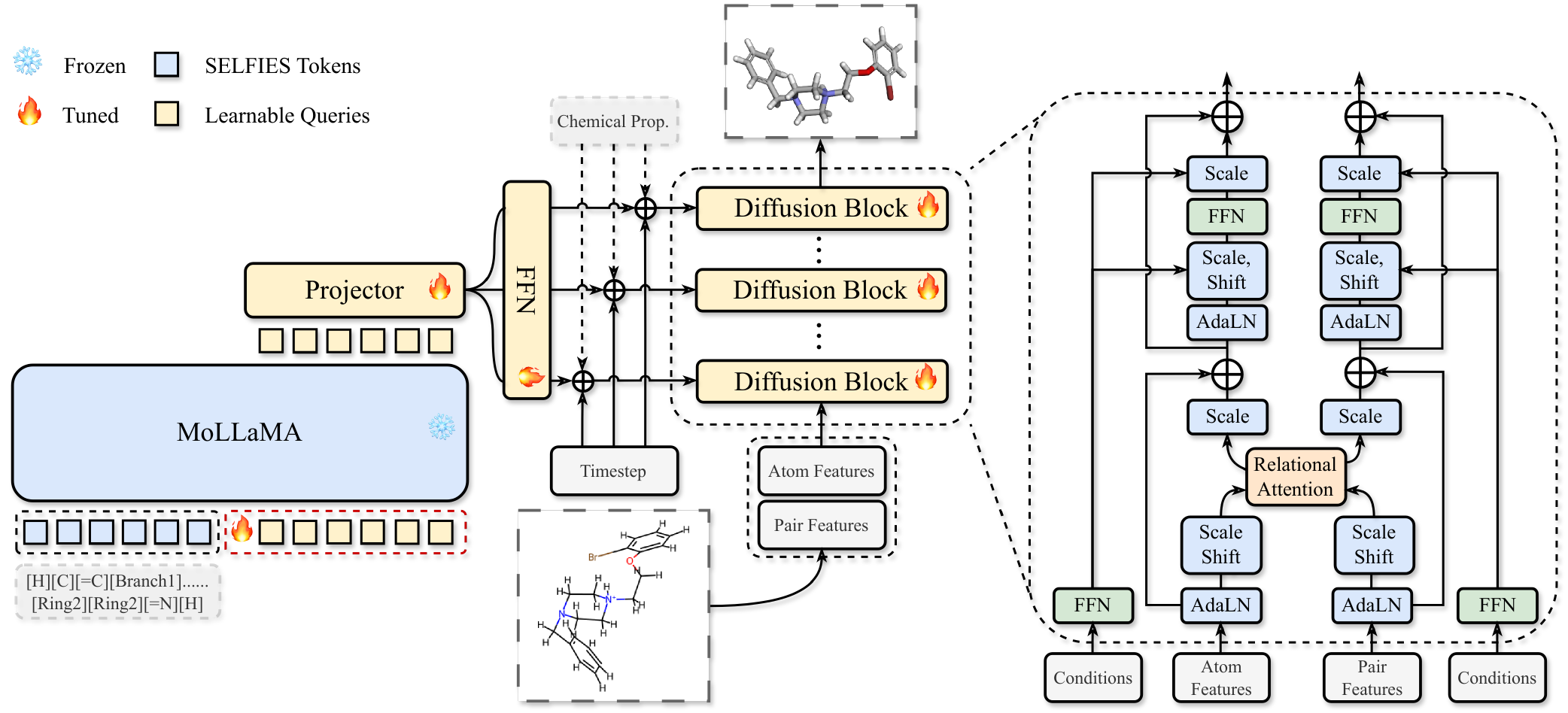}
    \caption{Overview of {\ours}. The model utilizes a set of learnable queries to probe a frozen, pre-trained molecular foundation model (MoLLaMA) and extract latent chemical knowledge. This condensed chemical priors, processed through a tunable projector and FFN, serves as a powerful condition to guide the subsequent generative process. A series of Diffusion Blocks, conditioned on this knowledge, timestep, and optional chemical properties, then generates the final 3D molecular geometries.}
    \label{fig:framework}
\end{figure*}
\subsection{Molecular Representation Learning}
Previous research extensively investigates multiple modalities for molecular representation, with a range of methods proposed based on the information leveraged during pretraining. SMILES-BERT~\cite{wang2019smiles} uses the SMILES sequence during pretraining to capture molecular representations. GEM~\cite{fang2022geometry} incorporates three-dimensional (3D) spatial structure information—atoms, bonds, and bond angles simultaneously—to model molecular representations. Chemformer~\cite{irwin2022chemformer} introduces a Transformer-based model that applies quickly to both sequence-to-sequence and discriminative cheminformatics tasks. The UniMol series~\cite{ji2024uni} evolves into a multimodal, molecular pretrained framework that integrates atomic-level, graph-level, and geometric-structure-level features to deliver comprehensive, context-aware understanding for chemistry and drug discovery. PRESTO~\cite{cao2024presto} further advances multimodal LLMs via cross-modal alignment and multi-graph understanding. Collectively, these developments demonstrate clear progress from single-modality sequence models to deeply integrated multimodal frameworks that align sequence, graph, and 3D geometric information.


\subsection{3D Molecule Generation}
Current research on 3D molecule generation traces its early foundations to autoregressive approaches \cite{gebauer2019symmetry,luo2022autoregressive}, which construct 3D molecular structures sequentially through the incremental addition of atoms or molecular fragments. Following the widespread success of diffusion models across diverse domains \cite{peebles2023scalable}, these models have emerged as the \textit{de facto} standard for 3D molecule generation \cite{hoogeboom2022equivariant, bao2022equivariant}. A critical limitation of early diffusion-based works, however, lies in their tendency to produce invalid molecules, a shortcoming stemming from their neglect of molecular bond information. To address this gap, subsequent diffusion-driven studies \cite{huang2024learning,vignac2023midi} incorporate bond generation by implementing a dedicated, separate diffusion process alongside backbone structure generation. Building on these advances, UDM-3D \cite{luo2025towards} elevates generation by conducting modeling within a unified latent space that integrates all molecular modalities, thereby enhancing both the efficiency of the generation pipeline and the quality of the resulting molecules. NExT-Mol \cite{liu2025next} improves 3D molecule generation by leveraging two key strengths of 1D SELFIES sequences: their inherent guarantee of 100\% molecular validity and their compatibility with more extensive training datasets~\cite{sterling2015zinc,tingle2023zinc}. Nevertheless, existing approaches either model only a single modality or heuristically fuse multiple modalities, without fully exploring deep cross-modal integration of chemical information. We therefore propose {\ours}, which utilizes learnable queries to extract inherent chemical knowledge from a 1D molecular foundation model and employs them as conditions for a diffusion model, thereby enhancing the performance of 3D molecular generation.
\section{Method}

\subsection{Preliminaries}
\paragraph{Geometric Graph Representation}
A molecule with $N$ atoms is represented as a geometric graph $\bm{G} = (\bm{A}, \bm{x}, \bm{h})$, where $\bm{x} = (\bm{x}^1, \ldots, \bm{x}^N) \in \mathbb{R}^{N \times 3}$ denotes the atomic coordinates that determine the molecular conformation. $\bm{h} = (\bm{h}^1, \ldots, \bm{h}^N) \in \mathbb{R}^{N \times d_1}$ represents atomic features, including a one-hot encoding of atom types and integer-valued formal charges. $\bm{A} \in \mathbb{R}^{N \times N \times d_2}$ encodes bond information; typically, one channel indicates bond existence, another indicates aromaticity, and a third specifies bond order. The subscript $t$ in $\bm{G}_t = (\bm{A}_t, \bm{x}_t, \bm{h}_t)$ denotes the time step in the diffusion process.

\paragraph{Molecular Diffusion Model}
In the forward diffusion process, Gaussian noise gradually perturbs the original coordinates $\bm{x}_{0}$:
\begin{equation}
q\!\left(\bm{x}_{t} \mid \bm{x}_{0}\right)
= \mathcal{N}\!\left(\bm{x}_{t};\,\sqrt{\bar{\alpha}_{t}}\,\bm{x}_{0},\,(1-\bar{\alpha}_{t})\,\bm{I}\right),
\end{equation}
where $t$ denotes the diffusion time step, and $\bar{\alpha}_{t}$ is a schedule hyperparameter controlling the noise scale at step $t$. By applying the reparameterization trick, we can sample:
\begin{equation}
\bm{x}_{t} = \sqrt{\bar{\alpha}_{t}}\,\bm{x}_{0} + \sqrt{1-\bar{\alpha}_{t}}\,\boldsymbol{\epsilon}_{t}, \quad
\boldsymbol{\epsilon}_{t} \sim \mathcal{N}(\bm{0}, \bm{I})
\end{equation}
Given the perturbed coordinates $\bm{x}^{(t)}$, the model is trained using mean-squared error (MSE) between the predicted noise and the ground truth sampled Gaussian noise $\boldsymbol{\epsilon}_{t}$:
\begin{equation}
\mathcal{L} = \big\|\boldsymbol{\epsilon}_{\theta}(\bm{A}_t, \bm{x}_t, \bm{h}_t,t) - \boldsymbol{\epsilon}_{t}\big\|_2^2
\end{equation}
In this work, the diffusion components primarily adopts the architectures of Diffusion Graph Transformer~\cite{huang2024learning} and Diffusion Molecule Transformer~\cite{liu2025next}, incorporating Relational Multi-head Self-attention (RMHA) and condition injection based on adaLN~\cite{peebles2023scalable} to propagate and update representations on geometric graphs.

\subsection{Transfer with Learnable Queries}
{\ours} introduces a set of randomly initialized learnable queries, denoted as $\bm{Q}$, to extract molecule-relevant conditions $\bm{C}$ from a frozen molecular foundation model. These conditions act as condensed chemical priors distilled from the foundation model’s inherent chemical knowledge and are projected into the input space of the diffusion model via a trainable projector. We fuse $\bm{C}$ with native conditioning (e.g., diffusion timestep and optional molecular property embeddings) through element-wise addition and train {\ours} end-to-end using the original diffusion objective, while keeping the foundation model frozen.

\paragraph{Learnable Queries}
We build on MoLLaMA~\cite{liu2025next}, a 960M-parameter molecular foundation model derived from LLaMA 2~\cite{touvron2023llama}, trained autoregressively on 1.8B molecules (90B SELFIES tokens) from ZINC-15~\cite{sterling2015zinc}. MoLLaMA encodes rich priors spanning: (1) 1D sequence chemistry: syntax and grammar of SELFIES, frequent motifs, functional groups, substructures, and distributional statistics. (2) 2D relational chemistry: graph-level dependencies implicitly learned via token co-occurrence, scaffold organization, and substituent patterns. To query out the chemical information that tightly couples to the model parameters, we attend a set number of learnable queries with $\bm{N}_Q$ tokens to the input tokens. This produces a compact, learnable bottleneck that adapts broad chemical knowledge to the downstream 3D generation without finetuning the backbone. Concretely, MoLLaMA acts as a powerful feature resampler, producing high-quality conditions for 3D molecule generation, while the queries interact with the SELFIES sequence and extract chemical signals most relevant to them.

\paragraph{Transfer from 1D to 3D}
We aim to transfer 1D sequence-level chemical latents in MoLLaMA to 3D molecular generative modeling. To achieve this, we introduce a projector based on the architecture of Qwen2.5~\cite{qwen2.5} transformer encoder. Unlike conventional language modeling, where queries are often ordered and sequential, our queries lack any inherent linguistic or sequential prior. We therefore enable bi-directional attention in the projector, allowing each query token to attend to all others and to the SELFIES context, capturing rich inter-dependencies and aggregating information comprehensively, which is formalized as:
\begin{equation}
[\bm{H}_i,\bm{Q}_i]=\text{Layer}_i(\bm{H}_{i-1},\bm{Q}_{i-1}),
\end{equation}
where $\bm{H}_i$ and $\bm{Q}_i$ denotes the SELFIES and queries representations in $i$th transformer encoder layer, respectively. $\text{Layer}_i$ represents the $i$th transformer encoder layer. After processing by the projector, the output representations of the queries encapsulate the distilled chemical priors tailored for the 3D molecule generation. Subsequently, we employ a Feed-forward Network (FFN) to map the projector’s output to the dimensionality required by the diffusion model’s conditioning interface. The strong conditions are injected into the diffusion model, guiding and enhancing the 3D molecular generation.

\subsection{Training Recipe}
We train {\ours} in two stages: first, we aim to produce high-quality 1D molecules that closely match the target data distributions, establishing a reliable foundation for subsequent 3D generation; concretely, we train for 100 epochs on QM9-2014 and 20 epochs on GEOM-DRUGS following \citet{liu2025next}. In the second stage, we use MoLLaMA in its pretrained form to provide a broader, more generalizable molecular prior, thereby avoiding the distribution sharpening associated with fine-tuning; MoLLaMA is kept frozen, while the projector, FFN, and diffusion model are trained end-to-end to optimize 3D structure generation based on the 1D molecules from stage 1.

By training {\ours} end-to-end, our approach enables effective transfer of 1D latents to the 3D generative space. This not only preserves the rich priors learned from large-scale molecular data, but also allows the diffusion model to adaptively utilize this knowledge for improved performance in 3D molecular generation.

\section{Experiments}
\begin{table*}[ht]
        \centering
        \caption{Performance of \textit{de novo} 3D molecule generation on GEOM-Drugs. Baselines are taken from~\citet{luo2025towards}. The best results in each metric are in \textbf{bold}.}
        \label{tab:denovo_geom_drugs}
        \setlength{\tabcolsep}{3mm}
        \resizebox{\linewidth}{!}{
        \begin{tabular}{lccccccccc}
            \toprule
            \textbf{2D-Metric} & \textbf{FCD}$\downarrow$ & \textbf{AtomStable} & \textbf{MolStable} & \textbf{V\&C} & \textbf{V\&U} & \textbf{V\&U\&N} & \textbf{SNN} & \textbf{Frag} & \textbf{Scaf} \\
            \midrule
            \bound{Train} & \bound{0.251} & \bound{1.000} & \bound{1.000}  & \bound{1.000} & \bound{1.000}  & \bound{0.000} & \bound{0.585}  & \bound{0.999} & \bound{0.584} \\
            CDGS & 22.051 & 0.991 & 0.706 & 0.285 & 0.285 & 0.285 & 0.262 & 0.789 & 0.022 \\
            JODO & 2.523 & 1.000 & 0.981 & 0.874 & 0.905 & 0.902 & 0.417 & 0.993 & 0.483 \\
            MiDi & 7.054 & 0.968 & 0.822 & 0.633 & 0.654 & 0.652 & 0.392 & 0.951 & 0.196 \\
            EQGAT-diff & 5.898 & \textbf{1.000} & 0.989 & 0.845 & 0.863 & 0.859 & 0.377 & 0.983 & 0.161 \\
            NExT-Mol & 0.334 & \textbf{1.000} & \textbf{0.999} & \textbf{1.000} & \textbf{0.999} & 0.945 & 0.529 & \textbf{0.999} & 0.552 \\
            UDM-3D & 0.692 & \textbf{1.000} & 0.925 & 0.879 & 0.913 & 0.907 & 0.525 & 0.990 & 0.540 \\
            \midrule
            \rowcolor{Green!10} \textbf{{\ours} (ours)} & \textbf{0.330} & \textbf{1.000} & \textbf{0.999} & \textbf{1.000} & \textbf{0.999} & \textbf{0.947} & \textbf{0.531} & \textbf{0.999} & \textbf{0.605} \\
            
            \midrule
            \textbf{3D-Metric} & \textbf{FCD$_\text{3D}$}$\downarrow$ & \textbf{AtomStable} & \textbf{MolStable}& \multicolumn{2}{c}{\textbf{Bond Length}$\downarrow$} & \multicolumn{2}{c}{\textbf{Bond Angle}$\downarrow$} & \multicolumn{2}{c}{\textbf{Dihedral Angle}$\downarrow$}  \\
            \midrule
                        
            \bound{Train} & \bound{13.73} & \bound{0.861} & \bound{0.028} & \multicolumn{2}{c}{\bound{1.56e-04}} & \multicolumn{2}{c}{\bound{1.81e-04}} &  \multicolumn{2}{c}{\bound{1.56e-04}} \\
            EDM & 31.29 & 0.831  & 0.002 & \multicolumn{2}{c}{4.29e-01} &\multicolumn{2}{c}{4.96e-01} &  \multicolumn{2}{c}{1.46e-02} \\
            JODO & 19.99 & 0.845 & 0.010 & \multicolumn{2}{c}{8.49e-02} &\multicolumn{2}{c}{1.15e-02} &  \multicolumn{2}{c}{6.68e-04}   \\
            MiDi & 23.14 & 0.750 & 0.003 & \multicolumn{2}{c}{1.17e-01} &\multicolumn{2}{c}{9.57e-02} &  \multicolumn{2}{c}{4.46e-03}  \\
            GeoLDM & 30.68  & 0.843 & 0.008 & \multicolumn{2}{c}{3.91e-01} &\multicolumn{2}{c}{4.22e-01} &  \multicolumn{2}{c}{1.69e-02} \\
            EQGAT-diff & 26.33 & 0.825 & 0.007 & \multicolumn{2}{c}{1.55e-01} &\multicolumn{2}{c}{5.21e-02} &  \multicolumn{2}{c}{2.10e-03}  \\
            NExT-Mol & 14.69 & 0.848 & - & \multicolumn{2}{c}{2.05e-02} &\multicolumn{2}{c}{8.18e-03} &  \multicolumn{2}{c}{2.31e-04}  \\
            UDM-3D & 17.36 & 0.852 & 0.014 & \multicolumn{2}{c}{\textbf{9.89e-03}} &\multicolumn{2}{c}{\textbf{5.11e-03}} &  \multicolumn{2}{c}{\textbf{1.74e-04}}\\
            \midrule
            \rowcolor{Green!10} \textbf{{\ours} (ours)} & \textbf{13.67} & \textbf{0.854}  & \textbf{0.026} & \multicolumn{2}{c}{7.87e-02} &\multicolumn{2}{c}{9.70e-03} &  \multicolumn{2}{c}{2.70e-04}  \\
            \bottomrule
        \end{tabular}
}
\end{table*}

\begin{table*}[ht]
        \centering
        \caption{Performance of \textit{de novo} 3D molecule generation on QM9. Baselines are taken from~\citet{luo2025towards}. The best results in each metric are in \textbf{bold}.}
        \label{tab:denovo_qm9}
        \setlength{\tabcolsep}{3mm}
        \resizebox{\linewidth}{!}{
        \begin{tabular}{lccccccccc}
            \toprule
            \textbf{2D-Metric} & \textbf{FCD}$\downarrow$ & \textbf{AtomStable} & \textbf{MolStable} & \textbf{V\&C} & \textbf{V\&U} & \textbf{V\&U\&N} & \textbf{SNN} & \textbf{Frag} & \textbf{Scaf} \\
            \midrule
            \bound{Train} & \bound{0.063} & \bound{0.999} & \bound{0.988}  & \bound{0.989} & \bound{0.989}  & \bound{0.000} & \bound{0.490}  & \bound{0.992}  & \bound{0.946} \\
            CDGS & 0.798 & 0.997 & 0.951 & 0.951 & 0.936 & 0.860 & 0.493 & 0.973 & 0.784 \\
            JODO & 0.138 & 0.999 & 0.988 & 0.990 & 0.960 & 0.780 & 0.522 & 0.986 & 0.934 \\
            MiDi & 0.187 & 0.998 & 0.976 & 0.980 & 0.954 & 0.769 & 0.501 & 0.979 & 0.882 \\
            EQGAT-diff & 2.088 & 0.999 & 0.971 & 0.965 & 0.950 & 0.891 & 0.482 & 0.950 & 0.703 \\
            SemlaFlow & 0.863 & 0.995 & 0.949 & 0.857 & 0.821 & 0.821 & 0.124 & -  & - \\
            NExT-Mol & 0.070 & \textbf{1.000} & 0.989 & \textbf{1.000} & 0.967 & 0.802 & 0.530  & \textbf{0.992} & \textbf{0.945} \\
            UDM-3D & 0.130 & 0.999 & 0.988 & 0.983 & \textbf{0.973} & \textbf{0.950} & 0.508 & 0.987 & 0.898 \\
            \midrule
            \rowcolor{Green!10} \textbf{{\ours} (ours)} & \textbf{0.065} & \textbf{1.000} & \textbf{0.999} & \textbf{1.000} & 0.964 & 0.804 & \textbf{0.533} & 0.991 & \textbf{0.945} \\
            
            \midrule
            \textbf{3D-Metric} & \textbf{FCD$_\text{3D}$}$\downarrow$ & \textbf{AtomStable} & \textbf{MolStable}& \multicolumn{2}{c}{\textbf{Bond Length}$\downarrow$} & \multicolumn{2}{c}{\textbf{Bond Angle}$\downarrow$} & \multicolumn{2}{c}{\textbf{Dihedral Angle}$\downarrow$}  \\
            \midrule
                        
            \bound{Train} & \bound{0.877} & \bound{0.994} & \bound{0.953} & \multicolumn{2}{c}{\bound{5.44e-04}} &\multicolumn{2}{c}{\bound{4.65e-04}} &  \multicolumn{2}{c}{\bound{1.78e-04}} \\
            EDM & 1.285 & 0.986 & 0.817 & \multicolumn{2}{c}{1.30e-01} &\multicolumn{2}{c}{1.82e-02} &  \multicolumn{2}{c}{6.64e-04} \\
            MDM & 4.861 & 0.992 & 0.896 & \multicolumn{2}{c}{2.74e-01} &\multicolumn{2}{c}{6.60e-02} &  \multicolumn{2}{c}{2.39e-02}   \\
            JODO & 0.885 & 0.992 & 0.934 & \multicolumn{2}{c}{1.48e-01} &\multicolumn{2}{c}{1.21e-02} &  \multicolumn{2}{c}{6.29e-04}   \\
            GeoLDM & 1.030 & 0.989 & 0.897 & \multicolumn{2}{c}{2.40e-01} &\multicolumn{2}{c}{1.00e-02} &  \multicolumn{2}{c}{6.59e-04} \\
            MiDi & 1.100 & 0.983 & 0.842 & \multicolumn{2}{c}{8.96e-01} &\multicolumn{2}{c}{2.08e-02} &  \multicolumn{2}{c}{8.14e-04}  \\
            EQGAT-diff & 1.520 & 0.988 & 0.888 & \multicolumn{2}{c}{4.21e-01} &\multicolumn{2}{c}{1.89e-02} &  \multicolumn{2}{c}{1.24e-03}  \\
            SemlaFlow & 1.127 & 0.971 & 0.787 & \multicolumn{2}{c}{-} &\multicolumn{2}{c}{-} &  \multicolumn{2}{c}{-}  \\
            ADiT & 2.884 & 0.211 & - & \multicolumn{2}{c}{9.98e-01} &\multicolumn{2}{c}{3.38e-02} &  \multicolumn{2}{c}{1.46e-03}  \\
            NExT-Mol & \textbf{0.879} & 0.993 & - & \multicolumn{2}{c}{\textbf{1.15e-01}} &\multicolumn{2}{c}{7.32e-03} &  \multicolumn{2}{c}{\textbf{1.95e-04}}  \\
            UDM-3D & 0.881 & 0.993 & 0.935 & \multicolumn{2}{c}{7.04e-02} &\multicolumn{2}{c}{9.84e-03} &  \multicolumn{2}{c}{3.47e-04}\\
            \midrule
            \rowcolor{Green!10} \textbf{{\ours} (ours)} & 0.952 & \textbf{0.995}  & \textbf{0.961} & \multicolumn{2}{c}{1.84e-01} &\multicolumn{2}{c}{\textbf{2.90e-03}} &  \multicolumn{2}{c}{3.08e-04}  \\
            \bottomrule
        \end{tabular}
}
\end{table*}

\begin{table*}[h]
        \centering
        \caption{Mean Absolute Error (MAE) for conditional 3D molecule generation on QM9. Baselines are taken from~\citet{luo2025towards}. The best results in each metric are in \textbf{bold}. $\Delta$ corresponds to the magnitude of the decrease in the MAE when comparing {\ours} to the runner-up method.}
        \label{tab:cond_qm9}
        \setlength{\tabcolsep}{5mm}
        \resizebox{\linewidth}{!}{
        \begin{tabular}{lcccccc}
            \toprule
            \textbf{Method} & $\mu(D)$ & $\alpha(Bohr^3)$ & $C_v(\frac{cal}{mol}K)$ & $\epsilon_{\text{HOMO}}(meV)$ & $\epsilon_{\text{LUMO}}(meV)$ & $\Delta\epsilon(meV)$ \\
            \midrule
            \bound{U-Bound} & \bound{1.613} & \bound{8.98} & \bound{6.879}  & \bound{645} & \bound{1457}  & \bound{1464}  \\
            \bound{L-Bound} & \bound{0.043} & \bound{0.09} & \bound{0.040} & \bound{39} & \bound{36} & \bound{65} \\
            EDM & 1.123 & 2.78 & 1.065 & 371 & 601 & 671 \\
            EEGSDE & 0.777 & 2.50 & 0.941 & 302 & 447 & 487 \\
            GeoLDM & 1.108 & 2.37 & 1.025 & 340 & 522 & 587 \\
            GeoBFN & 0.998 & 2.34 & 0.949 & 328 & 516 & 577 \\
            JODO & 0.628 & 1.42 & 0.581 & 226 & 256 & 335 \\
            NExT-Mol & 0.507 & 1.16 & 0.512  & 205 & 235 & 297 \\
            UDM-3D & 0.603 & 1.54 & 0.553 & 216 & 247 & 313 \\
            \midrule
            \rowcolor{Green!10} \textbf{{\ours} (ours)} & \textbf{0.338} & \textbf{0.29} & \textbf{0.131}  &\textbf{63} & \textbf{62}  & \textbf{85}  \\
            \textcolor{Green}{\textbf{$\Delta$}} & \textcolor{Green}{$\downarrow$\textbf{33.3\%}}
            & \textcolor{Green}{$\downarrow$\textbf{75.0\%}} & \textcolor{Green}{$\downarrow$\textbf{74.4\%}}  & \textcolor{Green}{$\downarrow$\textbf{69.3\%}} & \textcolor{Green}{$\downarrow$\textbf{73.6\%}}  & \textcolor{Green}{$\downarrow$\textbf{71.4\%}} \\
            \bottomrule
        \end{tabular}
}
\end{table*}

\subsection{Experimental Setup}
\paragraph{Datasets}
We train and evaluate {\ours} on GEOM-DRUGS~\cite{axelrod2022geom} and QM9-2014~\cite{ramakrishnan2014quantum}. QM9 is a standard small‑molecule set with bonding information, atom coordinates, and molecular properties for about 130K molecules with up to 9 heavy atoms (up to 29 atoms including hydrogen), containing single, double, and triple bonds. We split it into 100K/18K/13K for train/validation/test following~\cite{huang2024learning}. GEOM‑Drugs is larger, with molecules up to 181 atoms (average 44.4), 16 atom types, and bond types including single, double, triple, and aromatic. It provides multiple conformations with energies, from which we keep the lowest‑energy conformer per molecule, using an 8:1:1 train/validation/test split. All experiments are conducted on 4 NVIDIA A800-40GB GPUs. Experimental details are provided in the Appendix~\ref{apd:setting}.

\paragraph{Baselines}
We compare {\ours} with prior methods for \textit{de novo} 3D molecule generation, including CDGS~\cite{huang2023conditional}, JODO~\cite{huang2024learning}, MiDi~\cite{vignac2023midi}, EDM~\cite{hoogeboom2022equivariant}, MDM~\cite{huang2023mdm}, GeoLDM\cite{xu2023geometric}, EQGAT-diff~\cite{le2022representation}, SemlaFlow~\cite{irwin2024semlaflow}, ADiT~\cite{joshi2025all}, NExT-Mol~\cite{liu2025next}, and UDM-3D~\cite{luo2025towards}. For conditional generation, we additionally include EEGSDE~\cite{bao2022equivariant} and GeoBFN~\cite{song2024unified} as baselines. We sample 10K molecules from the training set three times and report average results as the performance upper bound.

\subsection{\textit{De Novo} 3D Molecule Generation}
\paragraph{Experimental Setting}
We adopt metrics from \cite{huang2024learning,hoogeboom2022equivariant} to ensure a comprehensive evaluation of structural validity (atom/bond features) and spatial arrangements (3D coordinates). The evaluation metrics fall into two groups:
\textbf{(1) 2D Metrics}: atom stability, validity \& completeness (V\&C), validity \& uniqueness (V\&U), validity \& uniqueness \& novelty (V\&U\&N), similarity to nearest neighbor (SNN), fragment similarity (Frag), scaffold similarity (Scaf), and Fréchet ChemNet Distance (FCD) \cite{polykovskiy2020molecular}.
\textbf{(2) 3D Metrics}: atom stability, FCD$_\text{3D}$ and maximum mean discrepancy (MMD) \cite{gretton2012kernel} for the distributions of bond length, bond angle and dihedral angle.

\paragraph{Results and Analysis} 
{\ours} establishes itself as a SOTA model for \textit{de novo} 3D molecule generation, demonstrating exceptional and well-rounded performance across both the complex, drug-like GEOM-Drugs dataset and the foundational QM9 dataset. As shown in the tables, {\ours} consistently achieves top-tier results, particularly in generating high-fidelity and stable 3D conformations.

As shown in Table~\ref{tab:denovo_geom_drugs}, {\ours}'s strength in modeling precise 3D geometries is evident. It achieves a SOTA FCD$_\text{3D}$ score of 13.67, outperforming all baselines and surpassing the upper bound (13.73). This indicates that the 3D feature distribution of molecules generated by {\ours} most closely resembles that of the real training data. Furthermore, {\ours} secures a SOTA 3D MolStable score of 0.026, confirming the superior physical realism and stability.

This high level of performance is mirrored on the QM9 dataset (Table~\ref{tab:denovo_qm9}). {\ours} achieves SOTA results in several key 2D metrics, including FCD (0.065), MolStable (0.999), V\&C (1.000), SNN (0.533), and Scaf (0.945), while also matching the best score for AtomStable (1.000). In 3D metrics, {\ours} achieves a SOTA Bond Angle MMD of 2.90e-03, significantly outperforming all other methods and demonstrating its superior ability to capture precise geometric details. Furthermore, it attains the best scores for 3D AtomStable (0.995) and 3D MolStable (0.961), underscoring its ability to produce conformations with the highest atomic and molecular stability among all models, surpassing the training data's stability profile. While its FCD$_\text{3D}$ score (0.952) is slightly behind the best (NExT-Mol at 0.879), it remains highly competitive.

In summary, {\ours}'s consistent SOTA performance across both benchmarks—leading in 3D fidelity and stability on GEOM-Drugs, and in bond angle precision and molecular stability on QM9—combined with its comprehensive excellence in 2D metrics, powerfully validates its capability to generate high-quality, geometrically precise, and stable 3D molecules for a wide range of applications.

\subsection{Conditional 3D Molecule Generation}
\paragraph{Experimental Setting}
To assess the model's capability for conditional molecular generation, we focus on a set of key quantum chemical properties. These properties include $C_v$, $\mu$, $\alpha$, $\epsilon_{\text{HOMO}}$, $\epsilon_{\text{LUMO}}$, and $\Delta\epsilon$. Following Fthe evaluation protocol described in \cite{hoogeboom2022equivariant,xu2023geometric}, we quantify the generation quality by calculating the Mean Absolute Error (MAE) between the properties of the generated molecules and their target values.

\paragraph{Results and Analysis}
Table 3 reports the Mean Absolute Error (MAE) for conditional 3D molecule generation on the QM9 dataset. Our method, {\ours}, achieves state-of-the-art performance by attaining the lowest MAE across all six evaluated properties, demonstrating its exceptional capability in generating molecules that precisely match target properties. Compared to the previous best-performing model, NExT-Mol, {\ours} shows substantial improvements across the board. Notably, it reduces the MAE for $\mu$ by 33.3\%, $\alpha$ by 75.0\%, $C_v$ by 74.4\%, $\epsilon_{\text{HOMO}}$ by 69.3\%, $\epsilon_{\text{LUMO}}$ by 73.6\%, and $\Delta\epsilon$ by a remarkable 71.4\%. These results underscore the effectiveness of our approach, which frames 3D molecule generation as "sculpting" 3D molecular geometries from chemical syntax. This allows {\ours} to effectively capture and reconstruct the intricate spatial and chemical information of molecules, leading to superior control and accuracy in property-conditional generation.

\begin{table*}[ht]
        \centering
        \caption{Ablation study on the architecture design. Atom., Mol., Len., and Ang. denote AtomStable, MolStable, Length, and Angle, respectively.}
        \label{tab:ablation}
        \setlength{\tabcolsep}{6mm}
        \resizebox{\linewidth}{!}{
        \begin{tabular}{lcccccc}
            \toprule
            \textbf{Design Choice} & \textbf{FCD$_\text{3D}$}$\downarrow$ & \textbf{Atom.} & \textbf{Mol.}  & \textbf{Bond Len.}$\downarrow$ & \textbf{Bond Ang.}$\downarrow$ & \textbf{Dihedral Ang.}$\downarrow$ \\
            \midrule
            \textbf{{\ours}} & \textbf{0.952} & \textbf{0.995} &  \textbf{0.961} & 1.84e-01 & \textbf{2.90e-03} & \textbf{3.08e-04}  \\
            Stage 2 \textit{w/o} MoLLaMA & 0.998  & 0.993 & 0.951 & \textbf{1.25e-01} & 7.31e-03 & 3.15e-04 \\
            Stage 2 \textit{w/} Finetuned MoLLaMA & 0.969 & 0.994 & 0.954  & 1.31e-01 & 2.90e-03 & 4.33e-04  \\
            \bottomrule
        \end{tabular}
}
\end{table*}

\begin{table}[ht]
        \centering
        \caption{Ablation study on the number of query tokens.}
        \label{tab:ablation_query}
        \setlength{\tabcolsep}{1mm}
        \resizebox{\linewidth}{!}{
        \begin{tabular}{ccccccc}
            \toprule
            \textbf{\# of Tokens} & \textbf{FCD$_\text{3D}$}$\downarrow$ & \textbf{Atom.} & \textbf{Mol.}  & \textbf{Bond Len.}$\downarrow$ & \textbf{Bond Ang.}$\downarrow$ & \textbf{Dihedral Ang.}$\downarrow$ \\ 
            \midrule
            16 & 0.972 & 0.994  & 0.955 & 1.76e-01 & \textbf{2.90e-03} & 3.48e-04 \\
            32 & 0.976 & 0.994 & 0.954  & \textbf{1.46e-01} & 3.10e-03 & 3.29e-04\\
            64 & \textbf{0.952} & \textbf{0.995} &  \textbf{0.961} & 1.84e-01 & \textbf{2.90e-03} & \textbf{3.08e-04} \\
            128 & 0.962 & \textbf{0.995} & 0.959 & 2.25e-01 & 3.40e-03 & 3.09e-04  \\
            256 & 0.964 & \textbf{0.995} & 0.958 & 1.47e-01 & 3.10e-03 & 3.57e-04 \\
            \bottomrule
        \end{tabular}
}
\end{table}

\subsection{Ablation Studies}
\paragraph{Training Strategy and Architecture Design}
We investigate the impact of our proposed architecture and training strategy, specifically the integration of the molecular foundation model, on \textit{de novo} 3D molecule generation. As shown in Table~\ref{tab:ablation}, we compare the full {\ours} against two variants: one without the MoLLaMA (``Stage 2 \textit{w/o} MoLLaMA") and one with finetuned MoLLaMA in stage 2 (``Stage 2 w/ Finetuned MoLLaMA"). The results indicate that removing the foundation model leads to a significant degradation in performance, with FCD$_\text{3D}$ increasing to 0.998 and a drop in molecular stability. This confirms that the 1D chemical priors extracted from MoLLaMA are crucial for guiding precise 3D generation. Furthermore, using the finetuned foundation model in stage 2 yields suboptimal results (FCD$_\text{3D}$ of 0.969) compared to keeping it frozen (0.952). This supports our hypothesis that freezing the pre-trained weights preserves the generalizable chemical knowledge and prevents the distribution sharpening or catastrophic forgetting that can occur when adapting to a smaller dataset. Consequently, our design choice of using a frozen foundation model with a learnable projector proves to be the most effective strategy for transferring latent chemical knowledge to 3D geometric space.

\paragraph{Query Tokens}
We further analyze how the number of learnable query tokens, $N_Q$, affects the model's ability to extract relevant chemical information. Table~\ref{tab:ablation_query} presents the performance variations as $N_Q$ ranges from 16 to 256. We observe that increasing the number of tokens from 16 to 64 leads to a consistent improvement in generation quality, with FCD$_\text{3D}$ decreasing from 0.972 to its optimal value of 0.952. This suggests that a sufficient number of queries is required to adequately capture the rich and complex chemical motifs encoded in the foundation model. However, further increasing the number of tokens to 128 and 256 results in a slight performance drop (e.g., FCD$_\text{3D}$ rises to 0.964). This degradation is likely due to the introduction of redundant information or noise, which complicates the optimization of the diffusion process. Therefore, we set $N_Q=64$ as the default configuration, striking an optimal balance between information sufficiency and representation compactness.

\begin{table}[t]
        \centering
        \caption{Ablation study on the number of projector layers.}
        \label{tab:ablation_layer}
        \setlength{\tabcolsep}{1mm}
        \resizebox{\linewidth}{!}{
        \begin{tabular}{ccccccc}
            \toprule
            \textbf{\# of Layers} & \textbf{FCD$_\text{3D}$}$\downarrow$ & \textbf{Atom.} & \textbf{Mol.}  & \textbf{Bond Len.}$\downarrow$ & \textbf{Bond Ang.}$\downarrow$ & \textbf{Dihedral Ang.}$\downarrow$ \\
            \midrule
            0 & 1.233  & 0.985  & 0.884 & 1.98e-01 & 6.00e-03 & \textbf{2.84e-04} \\
            6 & 0.977 & 0.994  & 0.952 & 1.67e-01 & \textbf{2.60e-03} & 3.79e-04  \\
            12 & 0.954 & \textbf{0.995} &  \textbf{0.961} & \textbf{1.49e-01} & 2.90e-03 & 2.90e-04 \\
            24 & \textbf{0.952} & \textbf{0.995} &  \textbf{0.961} & 1.84e-01 & 2.90e-03 & 3.08e-04 \\ 
            \bottomrule
        \end{tabular}
}
\end{table}
\paragraph{Projector Layers}
We examine the influence of the projector's depth on the cross-modal alignment between the 1D foundation model and the 3D diffusion generator. Table~\ref{tab:ablation_layer} compares projectors with 0, 6, 12, and 24 transformer layers. Using a projector with 0 layers results in the poorest performance, with a significantly high FCD$_\text{3D}$ of 1.233. This indicates that a simple FFN is insufficient to bridge the gap between the 1D semantic space and the 3D geometric space. Similarly, a shallow projector (6 layers) is less effective, yielding a higher FCD$_\text{3D}$ of 0.977. Increasing the depth to 12 layers significantly improves performance (FCD$_\text{3D}$ improves to 0.954), indicating that a deeper network is necessary to effectively map the semantic space of SELFIES to the geometric conditioning space. While increasing the layers to 24 yields the best overall performance (FCD$_\text{3D}$ of 0.952), the marginal gain over 12 layers suggests that performance begins to saturate. We adopt the 24-layer configuration to maximize the fidelity of the generated structures, ensuring the most accurate transfer of chemical syntax into 3D geometries.



\begin{figure*}[htbp]
    \centering 
    \begin{subfigure}[b]{0.48\textwidth}
        \centering
        \includegraphics[width=\textwidth]{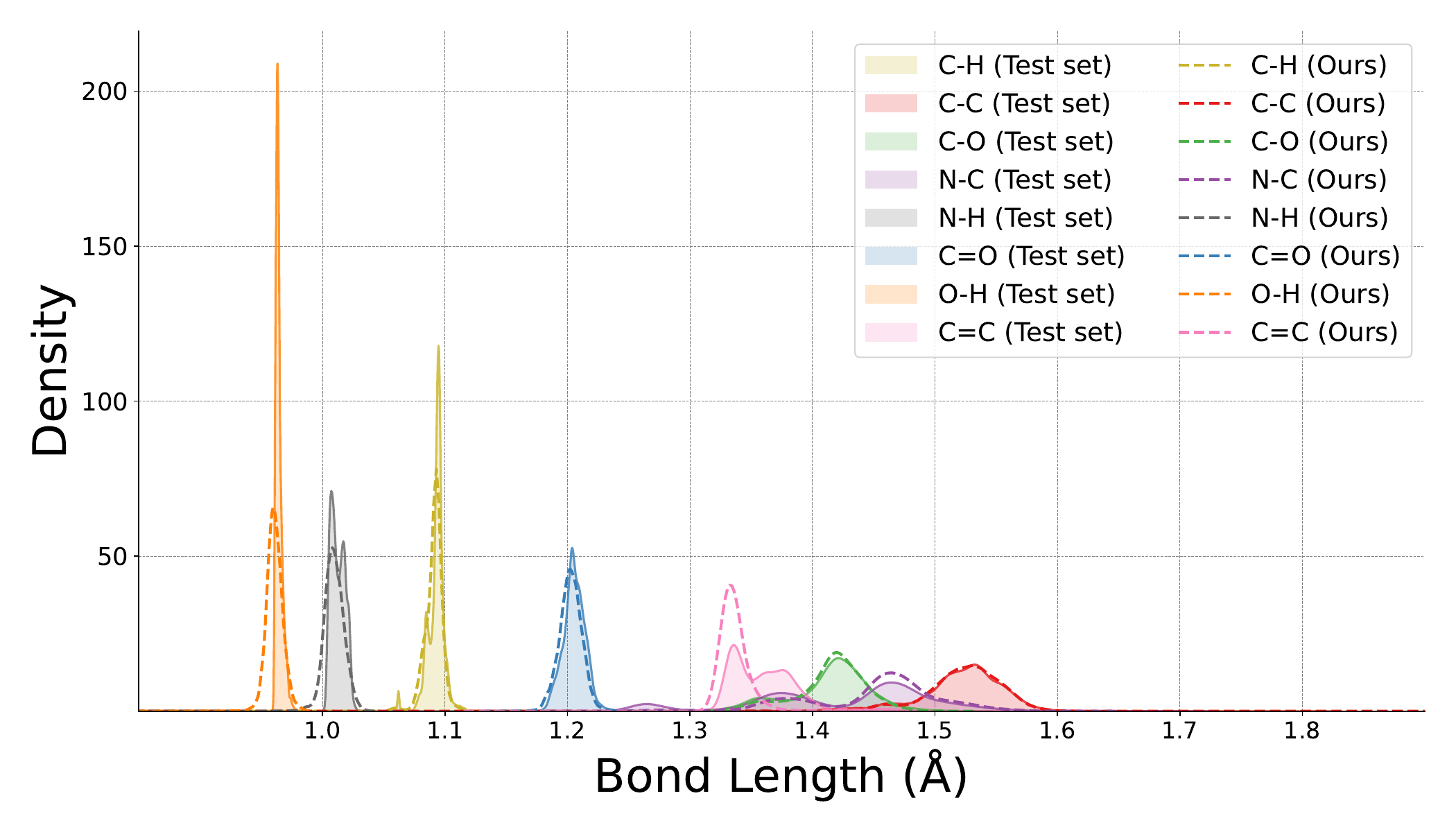} 
        \caption{QM9}
        \label{fig:distribution-a}
    \end{subfigure}
    \hfill 
    \begin{subfigure}[b]{0.48\textwidth}
        \centering
        \includegraphics[width=\textwidth]{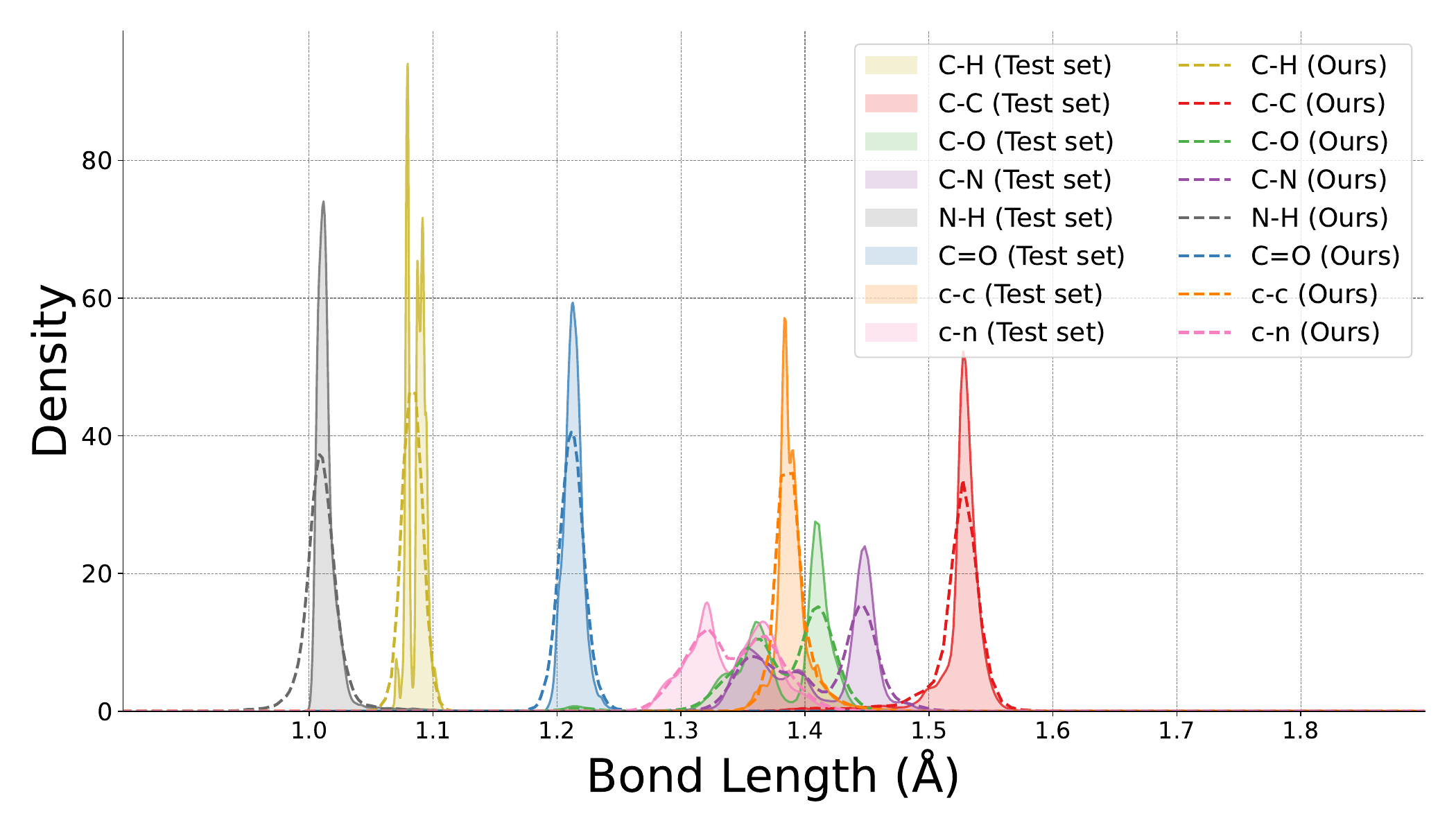} 
        \caption{GEOM-DRUGS}
        \label{fig:distribution-b}
    \end{subfigure}
    \caption{Distribution comparison of bond lengths between test set molecules and {\ours} generated molecules. Lowercase letters are used for atom types linked by aromatic bonds.}
    \label{fig:distribution}
\end{figure*}
\begin{figure*}[t]
    \centering
    \includegraphics[width=\linewidth]{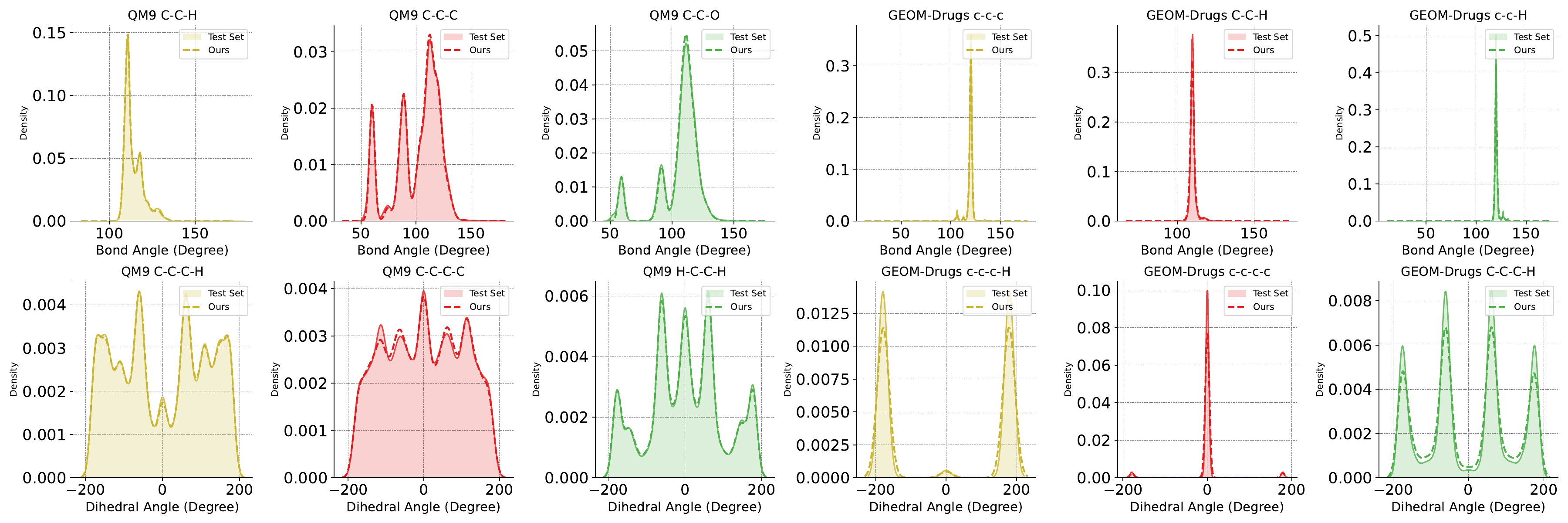}
    \caption{Distribution comparison of bond angle and dihedral angle between test set molecules and {\ours} generated molecules. Lowercase letters are used for atom types linked by aromatic bonds. Better view by zooming in.}
    \label{fig:distribution_angle}
\end{figure*}
\subsection{Further Analysis}
To evaluate the performance of {\ours} in generating geometrically accurate 3D structures, we compare the distributions of key geometric parameters—bond length, bond angle, and dihedral angle—between the generated samples and the test set, as shown in Figure~\ref{fig:distribution} and Figure~\ref{fig:distribution_angle}.

\paragraph{Distribution of Bond Length}
Figure~\ref{fig:distribution} displays the bond length distributions for several common bond types across the QM9 and GEOM-DRUGS datasets. {\ours} demonstrates a strong capability to capture the distinct distributions for different bond types (e.g., C-H, C-C, C=O). The peaks of the generated distributions align remarkably well with those of the test set (solid filled areas), indicating that the model has accurately learned the correct equilibrium bond length. The primary discrepancy appears in distributions with very sharp and high peaks, such as for C-H bonds, where the model's distribution is slightly broader and less peaked. This suggests the model imparts a minor degree of excess flexibility to these otherwise highly rigid bonds.

\paragraph{Distribution of Bond Angle and Dihedral Angle}
As shown in Figure~\ref{fig:distribution_angle}, the model's performance on bond angle and dihedral angle is excellent. It not only reproduces simple, unimodal distributions (e.g., the c-c-c angle in GEOM-Drugs) but also perfectly captures complex, multi-modal distributions (e.g., the C-C-O angle in QM9). Most impressively, the model excels at replicating the intricate, multi-peaked distributions of dihedral angle, such as for C-C-C-C in QM9 and c-c-c-H in GEOM-Drugs, which correspond to distinct, stable rotational conformers. 

In conclusion, the high fidelity of the generated distributions for bond length, bond angle, and dihedral angle provides strong evidence that {\ours} has successfully learned the fundamental geometric constraints of molecules, enabling it to generate chemically plausible and stable local geometries. We provide more distribution alignment figures of bond angle and dihedral angle in Appendix~\ref{apd:more}.

\subsection{Qualitative Analysis}
Figure~\ref{fig:geom_vis} and Figure~\ref{fig:qm9_vis} illustrate 3D molecules generated by our method. The results confirm that the generated molecules are chemically valid and possess diverse, complex structures, thereby highlighting the capability of {\ours} to produce realistic 3D molecular conformations.
\section{Conclusion}
In this work, we proposed {\ours}, a novel framework that effectively bridges the gap between 1D chemical syntax and 3D molecular geometry. By integrating a frozen 1D molecular foundation model (MoLLaMA) with a 3D diffusion model, {\ours} leverages learnable queries and a trainable projector to extract and transfer inherent chemical knowledge into the 3D generation process. This ``sculpting" mechanism ensures that the generated molecules not only benefit from the syntactic validity of SELFIES but also possess high-fidelity geometric structures. Extensive experiments on the GEOM-DRUGS and QM9 datasets demonstrate that {\ours} achieves state-of-the-art (SOTA) performance in both \textit{de novo} and conditional 3D molecule generation. Our model significantly outperforms existing baselines in terms of molecular stability, validity, and property-conditional accuracy (e.g., reducing MAE for key quantum properties). These results validate the effectiveness of cross-modal knowledge transfer in molecular representation learning. Future work may explore scaling this approach to larger biomolecules or integrating more diverse chemical modalities to further advance drug discovery.


\bibliography{references}
\bibliographystyle{icml2025}

\newpage
\appendix

\section{Detailed Experimental Settings}
\label{apd:setting}

\paragraph{Diffusion Model Configurations}
Table~\ref{tab:diffusion_hyp} presents the key hyperparameters used to train the diffusion model. Task-specific hyperparameters, such as batch size and training epochs, are detailed separately in the subsequent sections. We utilize the same atom and pair features as~\citet{jing2022torsional}. Specifically, the atom features consist of 74 and 44 dimensions for GEOM-DRUGS and QM9-2014, respectively, while the bond features consist of 4 dimensions.

\begin{table}[t]
\centering
\small
\caption{Hyperparameters of the diffusion model.}
\label{tab:diffusion_hyp}
\begin{tabular}{lcc} \toprule
                       &   \\ \midrule
n layers               & 10    \\
atom hidden size       & 512 \\
atom intermediate size & 2048 \\
pair hidden size       & 128 \\
pair intermediate size & 512 \\
n heads                & 8 \\
total params           & 55M \\
optimizer              & AdamW \\
init lr                & 1.00E-04 \\
min lr                 & 1.00E-05 \\
warmup lr              & 1.00E-06 \\
warmup steps           & 1000 \\
weight decay           & 0.05 \\
\bottomrule
\end{tabular}
\end{table}

\paragraph{\textit{De Novo} Molecule Generation}
For QM9-2014, we use a batch size of 512 and train for 100 epochs; for GEOM-DRUGS, we use a batch size of 256 and train for 20 epochs. During sampling, we employ a temperature of 1.0 and a beam size of 1.

We apply a dropout rate of 0.1 for QM9-2014 and 0.05 for GEOM-DRUGS. Following~\citet{huang2024learning}, we exclusively select the lowest-energy conformer for training on the GEOM-DRUGS dataset. For both datasets, we train {\ours} for 800 epochs, using batch sizes of 2048 for QM9-2014 and 512 for GEOM-DRUGS.


\paragraph{Conditional Molecule Generation}
Following NeXT-Mol~\cite{liu2025next}, we incorporate property-specific information into MoLLaMA using a condition MLP that encodes property data into a soft prompt. This soft prompt is prepended to the SELFIES input sequence embeddings prior to being fed into MoLLaMA. For the diffusion model, we employ an MLP to encode the property value, followed by a linear projection to align with the time embedding dimension. This processed condition is added to the time embedding, guiding the diffusion process toward the desired property throughout the denoising steps. We train MoLLaMA on the QM9-2014 dataset with a batch size of 256 for 100 epochs, using a sampling temperature of 1.0 and a beam size of 5. For the diffusion model, we train on QM9-2014 with a batch size of 512 for 800 epochs. During evaluation, we employ a dropout rate of 0 and utilize 100 sampling steps.

\section{More Results Analysis}
\label{apd:more}
\begin{figure*}[t]
    \centering
    \includegraphics[width=\linewidth]{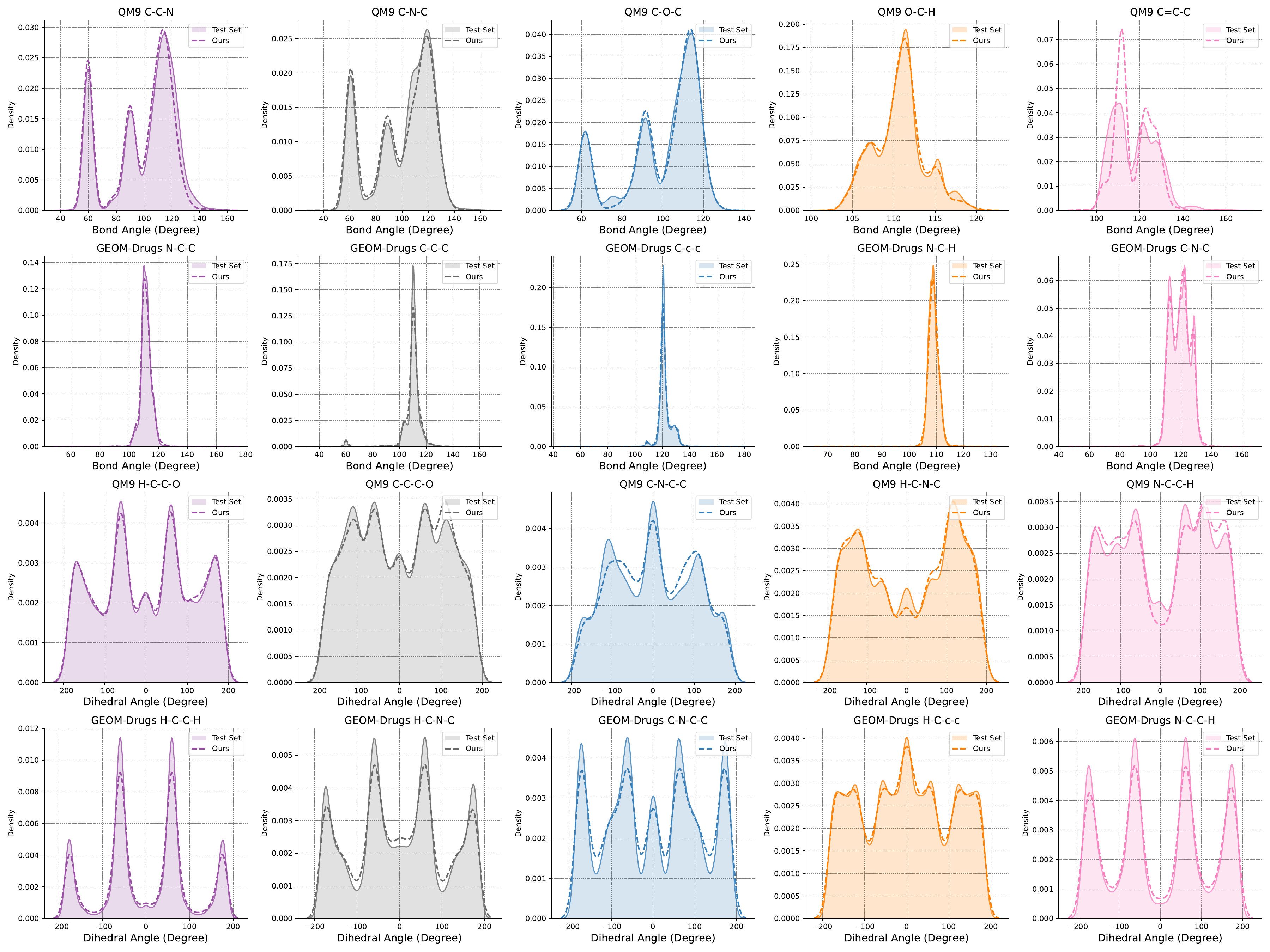}
    \caption{Extra distribution comparison of bond angles and dihedral angles between test set molecules and molecules generated from {\ours}. The corresponding substructures are reported in the titles.}
    \label{fig:more_dist}
\end{figure*}

In this section, we present additional experimental results to evaluate the generation quality of {\ours}. As illustrated in Figure \ref{fig:more_dist}, the distributions of bond angles and dihedral angles generated by {\ours} closely align with those of the test set, particularly for high-frequency substructures. This alignment demonstrates that {\ours} is capable of generating coherent chemically valid 3D geometries, rather than producing random spatial configurations.

\begin{figure*}[t]
    \centering
    \includegraphics[width=0.9\linewidth]{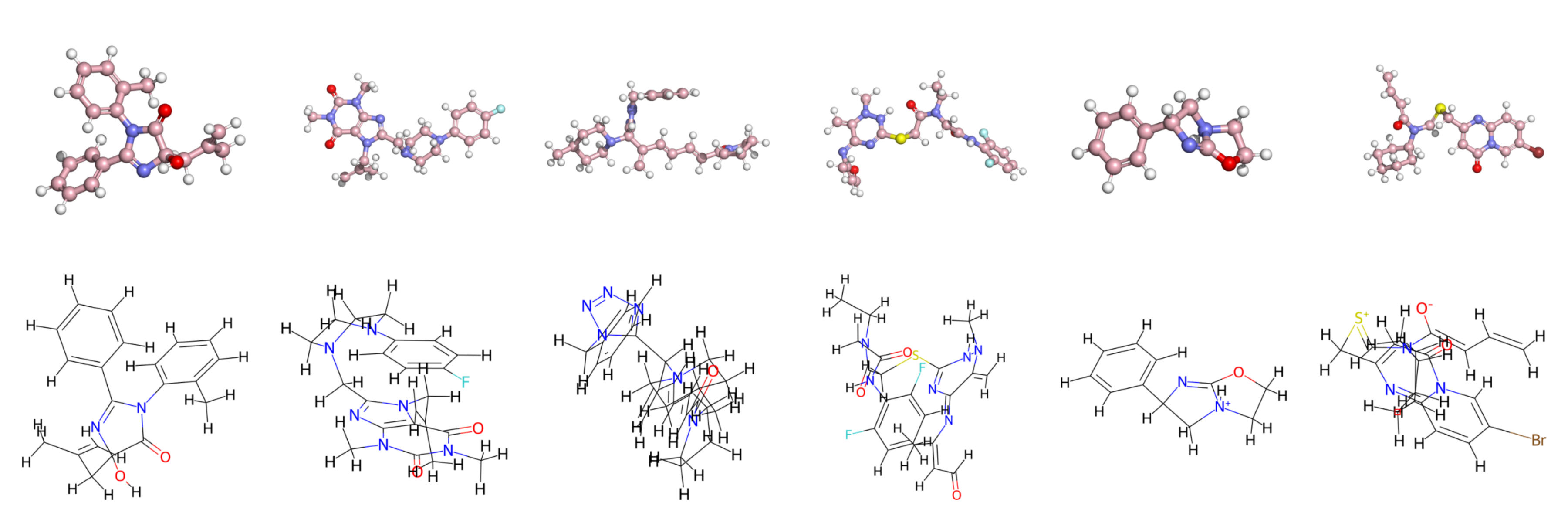}
    \includegraphics[width=0.9\linewidth]{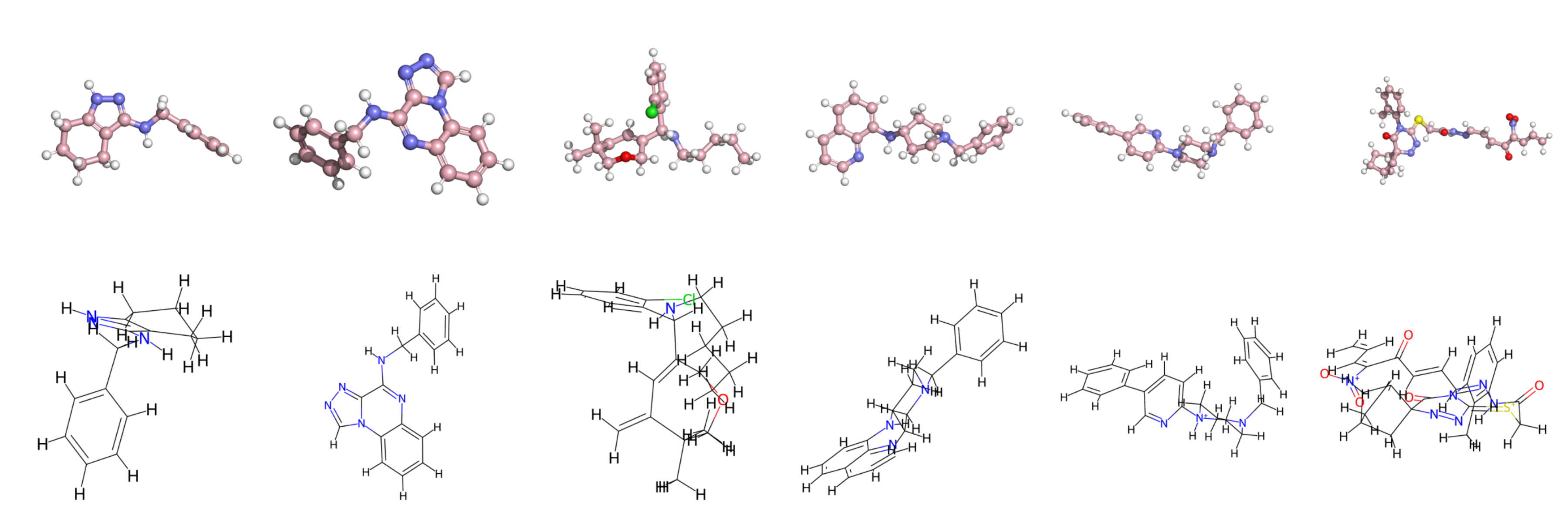}
    \includegraphics[width=0.9\linewidth]{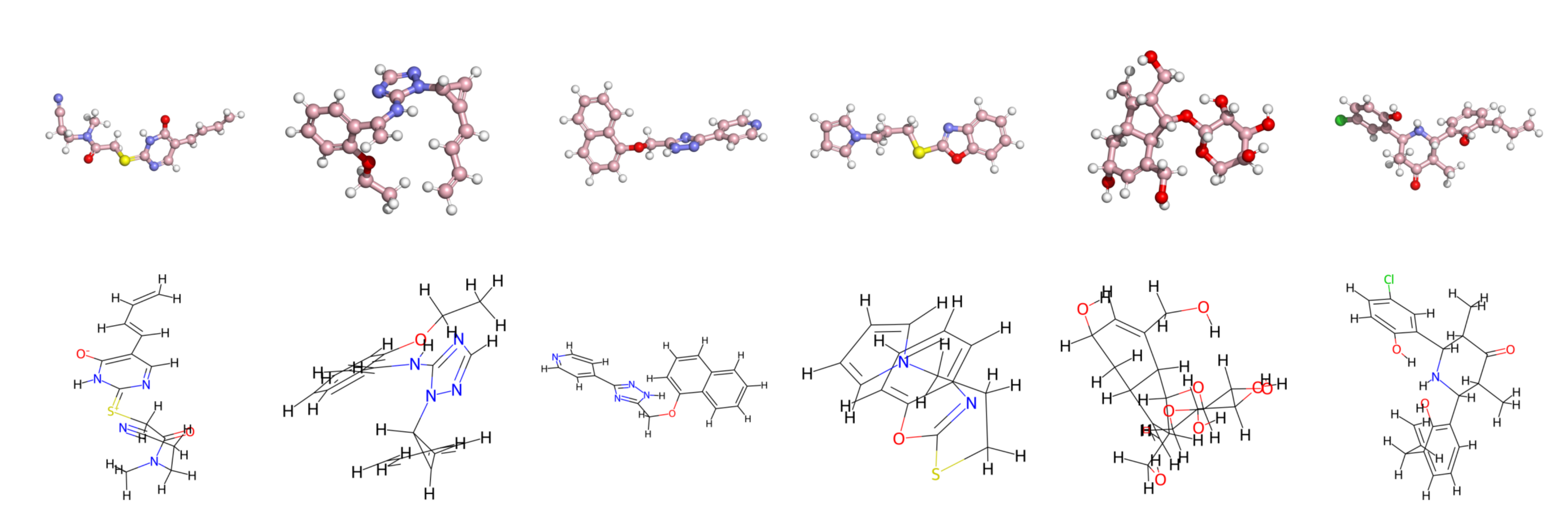}
    \includegraphics[width=0.9\linewidth]{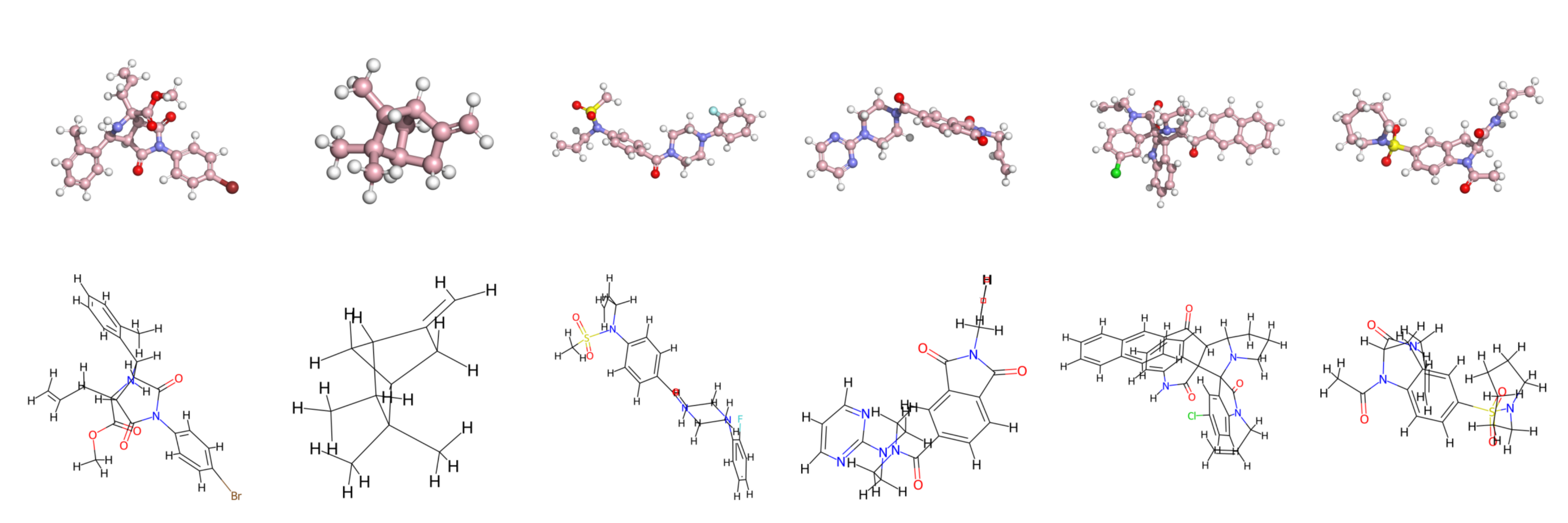}
    \caption{Visualization of random samples generated by {\ours} trained on GEOM-Drugs.}
    \label{fig:geom_vis}
\end{figure*}

\begin{figure*}[t]
    \centering
    \includegraphics[width=0.9\linewidth]{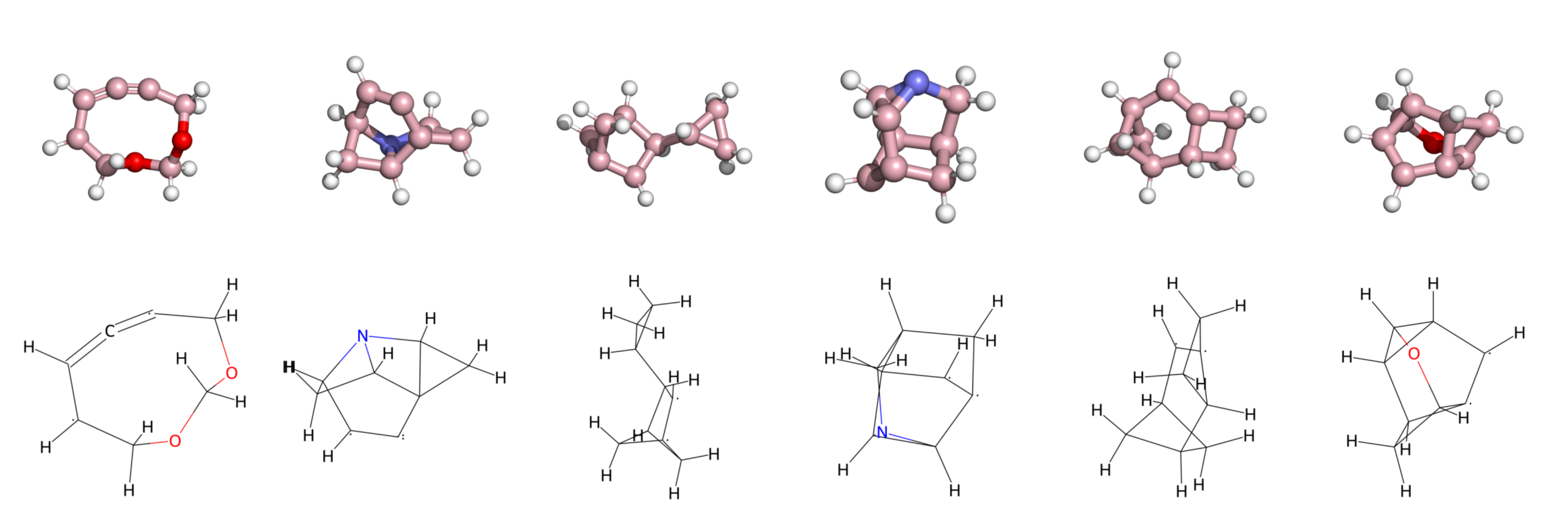}
    \includegraphics[width=0.9\linewidth]{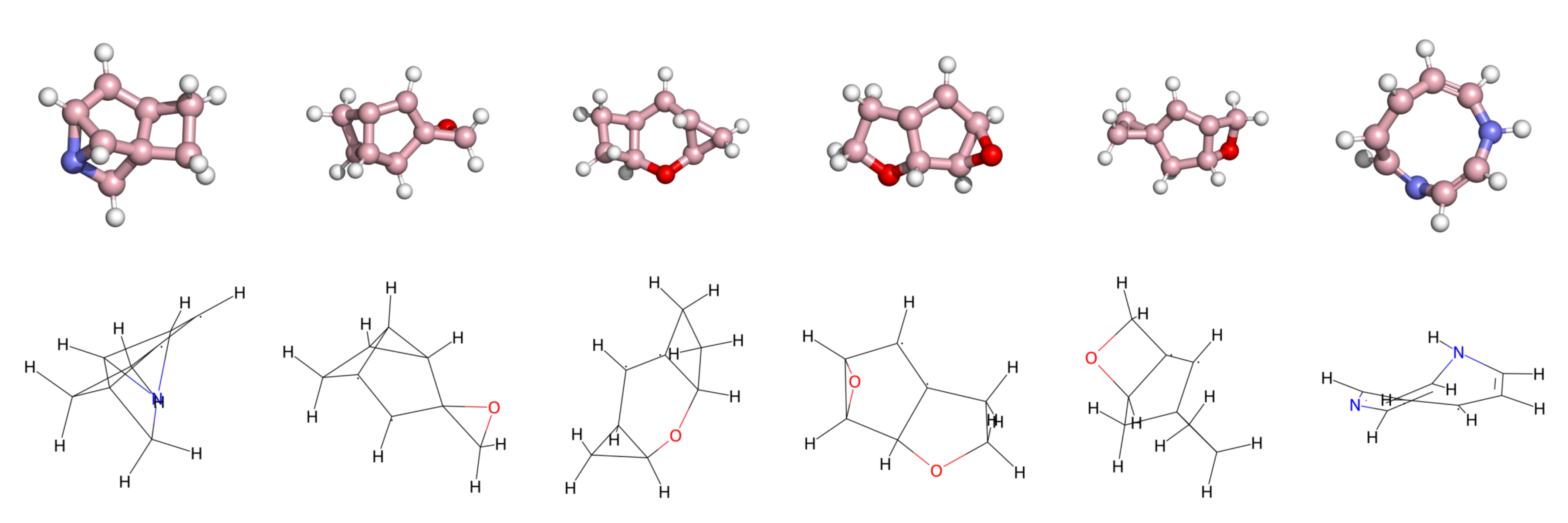}
    \includegraphics[width=0.9\linewidth]{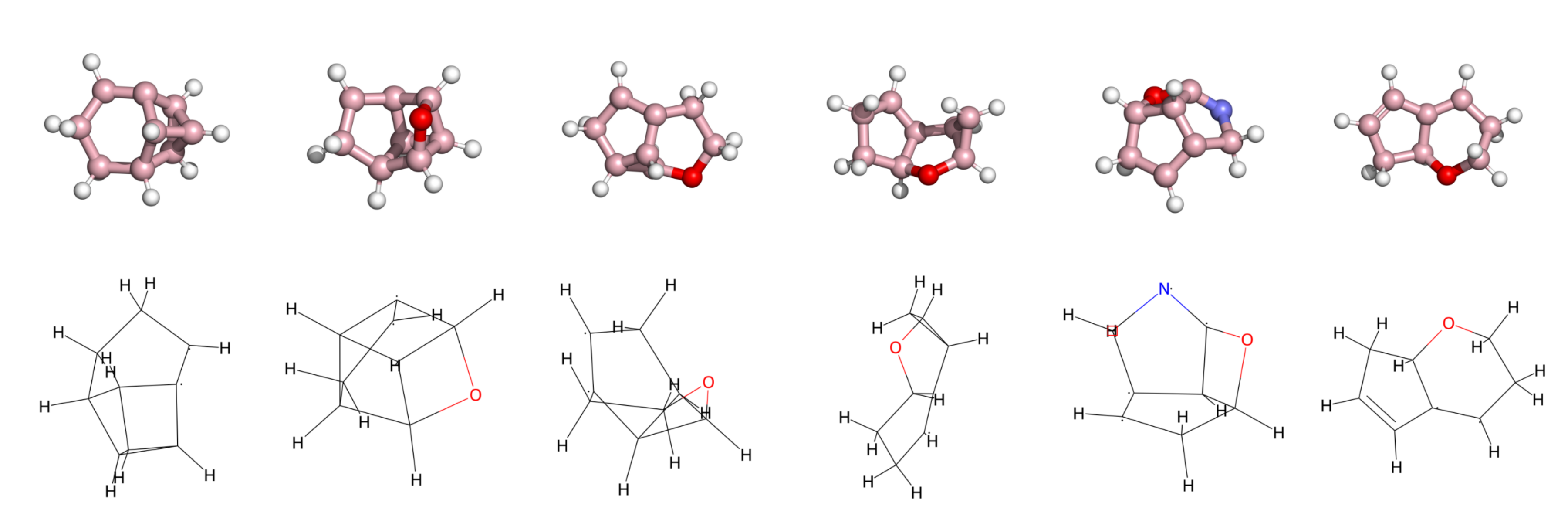}
    \includegraphics[width=0.9\linewidth]{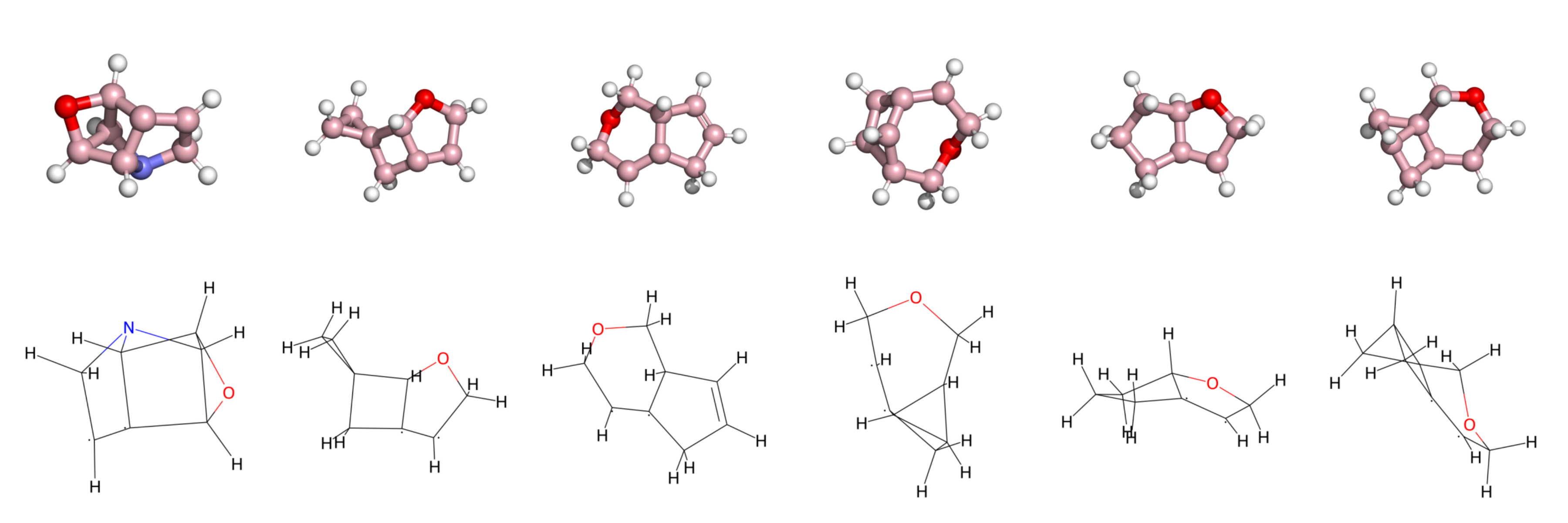}
    \caption{Visualization of random samples generated by {\ours} trained on QM9.}
    \label{fig:qm9_vis}
\end{figure*}

To further assess generation fidelity, we provide visualizations of complete molecules sampled from {\ours} trained on the GEOM-Drugs and QM9 datasets in Figure \ref{fig:geom_vis} and Figure \ref{fig:qm9_vis}, respectively. While the 3D geometry of certain conformers may be partially obscured by the viewing angle, the results confirm that {\ours} generates high-fidelity molecules with realistic 2D and 3D descriptors across a wide range of molecular weights. Notably, in the complex GEOM-Drugs dataset, {\ours} successfully avoids common generation artifacts such as disconnected components and effectively preserves the stable planarity of aromatic ring structures.


\end{document}